\documentclass[11pt]{article}
\usepackage[preprint]{acl}

\usepackage{times}
\usepackage{latexsym}
\usepackage[T1]{fontenc}
\usepackage[utf8]{inputenc}
\usepackage{microtype}
\usepackage{inconsolata}
\usepackage{hyperref}
\usepackage{url}
\usepackage{booktabs}
\usepackage{placeins}
\usepackage{float}
\usepackage{amsfonts}
\usepackage{nicefrac}
\usepackage{xcolor}
\usepackage{graphicx}
\usepackage{tikz}
\usetikzlibrary{arrows.meta}

\title{When Multiple Choices Are Correct:\\Asymmetric Probability Allocation in Jev and Other Decision Models}

\author{%
  Rushab Munot \\
  Independent Researcher \\
}

\begin{document}

\maketitle

\begin{abstract}
Decision models return normalized probability distributions over candidate choices, which downstream applications can use to select among choices. Although their providers emphasize the reliability of these probabilities, calibration on tasks with a single correct choice does not establish their reliability when multiple choices are correct. We evaluate three decision models (Jev, Liquid d1, and GPT-6 Luna Decisions) on three diverse datasets (MetaTool, HotpotQA, and ECtHR) to study probability allocation and thresholding. Even when models rank every correct choice above every distractor, in 28\% to 50\% of those lists, at least one correct choice receives a probability below the uniform probability of \(1/n\), where \(n\) is the number of choices. We also present each model with five identical copies of a correct choice and observe probability disparities, with mean gaps ranging from 17 to 79 percentage points. We further show how these allocation disparities affect the recovery of multiple correct choices under probability thresholding. Decision-model probabilities can rank correct choices, but a probability threshold does not identify the set of correct choices.

\end{abstract}

\section{Introduction}
\suppressfloats[t]
\begin{table}[t]
\caption{An example from HotpotQA asks which of the two tennis players is older, and lists 10 passages. The two passages below are necessary and sufficient together. However, the models concentrate almost all probability on one passage. The full list of passages is in Appendix~\ref{app:example}.}
\label{tab:example}
\centering
\small
\begin{tabular}{@{}p{\linewidth}@{}}
\toprule
\textbf{Question:} who is older, Maria Sharapova or Chanda Rubin? \\
\textbf{Instruction:} Choose every passage in the list that supports the answer to the question. \\
\midrule
Chanda Rubin (\textbf{born February 18, 1976}) is an American tennis \ldots \\
Maria Yuryevna Sharapova (\ldots{} \textbf{born April 19, 1987}) is a Russian professional \ldots \\
\bottomrule
\end{tabular}

\setlength{\tabcolsep}{0pt}
\begin{tabular*}{\linewidth}{@{}l@{\extracolsep{\fill}}rrrrrr@{}}
 & \multicolumn{3}{c}{10 passage list} & \multicolumn{3}{c}{Yes/no} \\
\cmidrule{2-4} \cmidrule{5-7}
Passage & Jev & d1 & Luna & Jev & d1 & Luna \\
\midrule
Chanda Rubin    & 0.93 & 0.94 & 0.93 & 0.31 & 0.16 & 0.47 \\
Maria Sharapova & 0.07 & 0.06 & 0.02 & 0.14 & 0.07 & 0.00 \\
\bottomrule
\end{tabular*}

\end{table}

Decision models are emerging as an alternative to generative language models for applications requiring structured decisions, offering lower inference costs and latency by returning predictions directly rather than generating text \citep{almeida2026jev,liquidAI2026d1}. Three recently introduced models are TypeSafe AI's Jev \citep{almeida2026jev}, Liquid AI's d1 \citep{liquidAI2026d1}, and OpenAI's GPT-6 Luna Decisions \citep{openai2026decisions}. Despite differences in their implementations, these models expose a similar choice-based interface. Given contextual information, instructions, and a set of candidate choices, they return a probability distribution over the choices, with probabilities summing to one. Their providers emphasize the reliability and practical utility of these outputs. TypeSafe states that "Higher probability should correspond to a greater chance that the answer is correct" \citep{typesafe2026primer}, while Liquid AI describes its decision models as returning "calibrated probabilities across a fixed set of outcomes" \citep{liquid2026models}. OpenAI recommends setting probability thresholds for downstream routing, filtering, and review \citep{openai2026decisions}, and TypeSafe demonstrates using such thresholds to select additional correct choices beyond the highest-ranked choice \citep{typesafe2026choice}. These claims and recommendations raise a question for tasks with multiple correct choices. Can normalized Choice probabilities reliably identify all correct choices, rather than only the highest-ranked one?

While categorical distributions must output probabilities that sum to 1, providers document that these are calibrated and recommend thresholds in practice. Since probabilities are a share of the total of 1, with more correct choices, the share of each correct choice goes down on average. However, since thresholding is the recommended way to use these models, for multiple correct choices, each correct choice must have a baseline probability. The constraint still allows an even split. Two correct choices in a list of 10 can each receive 0.5, and five identical copies can each receive 0.2. We hence study the allocation of unit probability over choices passed to the model.

Consider the HotpotQA question ``Who is older, Maria Sharapova or Chanda Rubin?'' \citep{yang2018hotpotqa}. The question is accompanied by 10 candidate passages. One of those passages gives Rubin's date of birth and another gives Sharapova's. Both passages together are sufficient, but neither is sufficient alone. However, all three decision models assign a probability higher than 0.90 to Rubin's passage and a probability lower than 0.10 to Sharapova's passage. The Choice probabilities therefore concentrate on one of the two passages needed to answer the question. Table~\ref{tab:example} shows the question, the two relevant passages and the probabilities. After replacing the names, tournaments, dates and locations with invented ones, the Choice probabilities remain uneven. Jev and Liquid d1 still favor the older player's passage, but GPT-6 Luna Decisions now favors the younger player's passage. Appendix~\ref{app:example} lists the full set of original and anonymized passages, along with the probabilities assigned by each model.

Recent work on decision models has measured accuracy and calibration on datasets with one correct choice~\citep{porcedda2026jev}. \citet{deusser2026evaluating} evaluate Jev on 37 datasets and report that its Choice probabilities are calibrated on their tasks. They also find that rotating choices leaves aggregate accuracy unchanged. On yes/no questions, Jev's probabilities rank well but are poorly placed relative to a threshold of 0.5. \citet{liu2026dice} finds that Jev gives unequal probabilities to the six faces of a fair die. Both studies test Jev only. \citet{li2026beyond} compare Jev's answers to logically related yes/no and Choice questions and find differences between probabilities for events that should agree. TypeSafe states that the order of choices can affect Jev's answer \citep{typesafe2026jev}. \citet{li2026readout} find that restricting and renormalizing the original probabilities predicts aggregate accuracy after choices are removed. For generative language models, \citet{zheng2024large,pezeshkpour2024large} show that choice order changes answers to multiple-choice questions. \citet{blankenstein2026biasbusters} find that language models selecting among functionally equivalent tools can favour particular providers. \citet{wang2025leastincorrect} find that LLMs can select an offered answer even when none is correct. However, modern AI tasks require selecting more than one correct choice. Examples include selecting the tools an agent needs to complete a task, the diagnoses supported by a medical record, the passages relevant to a query in retrieval-augmented generation, and the vulnerability types present in code.

In this paper, we measure how decision models distribute probability over several correct choices. We specifically examine the gap between the maximum and minimum probabilities assigned to correct choices. We first evaluate the models on the choice lists in the datasets. We then modify the datasets' choice lists while keeping the query and the instruction fixed to present five identical copies of a correct choice. Identical copies differ only in their place in the list and their random IDs, so they should get the same probability. We also check calibration on yes/no questions and observe miscalibration on intermediate probabilities. We study their behavior on three datasets, MetaTool \citep{huang2024metatool}, ECtHR \citep{chalkidis2022lexglue}, and HotpotQA \citep{yang2018hotpotqa}.

Our results show that decision-model probabilities are not reliable for identifying multiple correct choices. When a list contains more than one correct choice, the mean gap between the highest and the lowest probabilities assigned to correct choices ranges from 55 to 74 percentage points across the three models and three datasets. A uniform distribution over \(n\) choices assigns \(1/n\) to each, representing no preference among them. The models rank every correct choice above every distractor in 52\% to 85\% of the lists. In 28\% to 50\% of those lists, a correct choice receives less than \(1/n\). Further, even when the models rank every correct choice above every distractor, the best single threshold gives the exact set in only 64\% to 78\% of those examples, and that threshold is low, ranging between 0.02 and 0.08, against the 0.25 that TypeSafe uses as an example in its documentation (Section~\ref{sec:results}).

The allocation of probabilities is uneven even when a list contains five identical copies of a correct choice. The mean gap between the highest and lowest (identical) copies ranges from 17 to 79 percentage points.
For lists containing exactly one correct choice, Jev's ECE is 8 percentage points on MetaTool and 3 on ECtHR, consistent with \citet{deusser2026evaluating}. When multiple choices are correct, normalized Choice probabilities cannot be interpreted as independent probabilities of correctness. On yes/no questions about each choice, the ECE is 3 to 9 percentage points over all bins and 8 to 18 percentage points without the bin from 0 to 0.1.

\section{Experimental Setup}
\label{sec:setup}

\begin{table}[t]
\caption{The three datasets, with the number of examples by their number of correct choices, given in brackets for two or more. A list has 10 choices, except the MetaTool scenario lists, with 5, 10 or 15 tools (100, 1,400 and 300 queries), and 60 HotpotQA lists, with 2 to 9 passages.}
\label{tab:datasets}
\centering
\footnotesize
\setlength{\tabcolsep}{4pt}
\begin{tabular}{llll}
\toprule
 & MetaTool & HotpotQA & ECtHR \\
\midrule
Query & Query & Question & Case facts \\
Choice & Tool & Passage & Article \\
Choices per list & 5 to 15 & 2 to 10 & 10 \\
One correct choice & 2,795 & -- & 655 \\
Two or more & 497 (2) & 7,405 (2) & 192 (2 to 4) \\
No correct choice & 995 & -- & 153 \\
\bottomrule
\end{tabular}
\end{table}

\paragraph{Models.}
We study three decision models. We call Jev (\texttt{jev-1.13.0})~\citep{almeida2026jev} through TypeSafe's API, and Liquid d1 (\texttt{liquid/d1})~\citep{liquidAI2026d1} and GPT-6 Luna Decisions (\texttt{openai/gpt-6-luna-decisions})~\citep{openai2026decisions} through Vercel AI Gateway~\citep{vercel2026models}. Liquid AI and OpenAI publish no version for these two models, and our results refer to the models that the gateway served in October 2026. A request holds a query, an instruction and a list of named choices, each with a description. The answer holds the selected choice, which we call the prediction, and one probability per choice. These Choice probabilities sum to 1~\citep{typesafe2026choice}. We call the highest probability of an answer its confidence. This is not the \texttt{confidence} field of Jev's answer, which TypeSafe computes from the highest probability and the number of choices~\citep{typesafe2026confidence}. Jev and GPT-6 Luna Decisions return probabilities with two decimals, and Liquid d1 with full precision. GPT-6 Luna Decisions can refuse a question and return no probabilities. Jev and Liquid d1 refused none. The three models received the same requests, and we sent each request once. Table~\ref{tab:runs} gives the number of examples and lists of each kind.

\paragraph{Datasets.}
We chose three datasets from different domains, all of which allow multiple correct choices, or have a subset of examples that do. MetaTool~\citep{huang2024metatool} is a tool selection task. The model sees the query and a list of tools and is asked to select the tools needed to solve the query. Its four tool selection subsets hold 4,287 queries. Of these, 995 have one correct tool among tools with similar descriptions, 1,800 have one correct tool among the tools of a scenario, 995 have no correct tool, and 497 have two correct tools. Some queries appear in more than one subset, and the answer to a repeated request counts once for each example.

The distractor setting of HotpotQA~\citep{yang2018hotpotqa} gives a query and up to ten passages, two of which support the answer. While HotpotQA contains training, validation and test sets, we only use the validation set in this study.
 ECtHR, Task~A of LexGLUE~\citep{chalkidis2022lexglue}, is a legal multi-label classification task. Each example
 gives the facts of a case and ten articles of the Convention and it asks the model to select the articles that the case violates. Of its 1,000 test cases, 655 have one violated article, 192 have two to four, and 153 have none. When several choices are correct, the instruction asks for all of them, for example ``Choose every passage in the list that supports the answer to the question.'' Table~\ref{tab:datasets} gives the number of examples of each kind of list.

\paragraph{Identical choices.}
In our first experiment, for every example, we send five identical copies of one of its correct choices. The query and the instruction remain unchanged. Exactly identical copies are rejected by the APIs, hence we name each copy with a random ID. Identical copies are equally correct and equally relevant to the query, so a difference between their probabilities cannot come from their content. To place every copy once at every position, we send each list in its five cyclic permutations~\citep{zheng2024large,pezeshkpour2024large}.

We build these lists for 3,292 MetaTool queries, a seeded sample of 1,000 HotpotQA questions and the 847 ECtHR cases with a violated article, 25,695 lists per model. We report the mean probability of the highest and the lowest copy, the gap between them, and the share of predictions at each position. Under position-invariance and uniformity, we would expect the mean probability of each copy to be 0.20, and the gap to be 0.00.

\paragraph{The gap between correct choices.}
Next, we send the lists as the datasets give them. MetaTool gives 497 queries with two correct tools, and HotpotQA its questions. We send both in their released order. ECtHR gives 192 cases with two to four violated articles. We send each list in its ten cyclic permutations. For each answer, the gap is the highest minus the lowest probability given to a correct choice, and we report its mean. As a reference we use the uniform distribution, which gives each of $n$ choices $1/n$ and prefers none of them. We report the share of lists in which at least one correct choice gets less than $1/n$. To separate this from ranking errors, we also report the share of lists in which every correct choice is above every distractor, and, among those lists, the share in which a correct choice still gets less than $1/n$. These counts use only the released order of the ECtHR lists and leave out the 21 HotpotQA questions whose list has no distractor.

\paragraph{Calibration.}
A model is calibrated when, among the answers to which it gives a probability $p$, a share $p$ is correct. For lists with exactly one correct choice, we define confidence as the highest probability assigned to any choice. For lists with several correct choices, this definition does not directly apply. We therefore convert each list into several yes/no questions, one per choice, based on the benchmarks' format. For example, for HotpotQA we ask ``Does this passage support the answer to the question?'' for every (question, passage) pair. The models respond with a single probability of ``yes''. We report the Expected Calibration Error (ECE)~\citep{naeini2015obtaining,guo2017calibration} over ten bins of equal width, which compares the mean confidence, or the mean probability of ``yes'', with the observed share of correct predictions, or of true ``yes'' labels, in each bin. For the yes/no questions, we also report the ECE without the bin from 0 to 0.1.

\paragraph{Selecting by a threshold.}
Model providers suggest using a threshold~\citep{openai2026decisions}, and TypeSafe uses one to select multiple correct choices~\citep{typesafe2026choice}. For example, TypeSafe's documentation includes an example with three choices, two of which are correct, and uses a threshold of 0.25. A threshold $t$ selects every choice whose probability is at or above $t$. In this experiment, we study whether there is a common threshold that selects all correct choices and no distractors across all examples in the dataset.

We start with examples answered both as lists and as yes/no questions, vary $t$ from 0 to 1, and compare the two formats. We define exact set accuracy as the share of examples in which every correct choice is selected and every distractor is rejected. We define recall as the proportion of correct choices above the threshold, and precision as the proportion of choices above the threshold that are correct. We compute precision, recall, and F1 from counts pooled across all examples. We plot exact set accuracy as a function of the threshold $t$ from 0 to 1, and precision, recall and F1 are reported in Appendix~\ref{app:thresholds}.

\paragraph{None of the above.}
Following \citet{tam2025none}, we send lists with no correct choice (Appendix~\ref{app:abstention}).

\section{Results}
\label{sec:results}

\begin{table*}[t]
\caption{For lists with several correct choices, \emph{Gap} is the mean probability of the highest correct choice minus that of the lowest, in percentage points, \emph{Ranked} is the share of lists with every correct choice above every distractor, and \emph{Below $1/n$} is the share of those lists with a correct choice below 1/n for a list of n choices. $1/n = 0.1$ for every list except 39 HotpotQA lists of 3 to 9 passages. The two shares use the released order of the ECtHR lists and leave out the lists with no distractor.}
\label{tab:allocation}
\centering
\footnotesize
\begin{tabular}{lrrrrrrrrr}
\toprule
 & \multicolumn{3}{c}{MetaTool, 2 correct} & \multicolumn{3}{c}{HotpotQA, 2 correct} & \multicolumn{3}{c}{ECtHR, 2 to 4 correct} \\
\cmidrule(lr){2-4} \cmidrule(lr){5-7} \cmidrule(lr){8-10}
Model & Gap & Ranked & Below $1/n$ & Gap & Ranked & Below $1/n$ & Gap & Ranked & Below $1/n$ \\
\midrule
Jev & 74 & 73\% & 50\% & 60 & 82\% & 31\% & 58 & 79\% & 36\% \\
Liquid d1 & 55 & 70\% & 29\% & 55 & 85\% & 28\% & 57 & 76\% & 42\% \\
GPT-6 Luna Decisions & 55 & 52\% & 35\% & 57 & 70\% & 29\% & 61 & 73\% & 45\% \\
\bottomrule
\end{tabular}

\end{table*}

\begin{table}[t]
\caption{Each list has 5 copies of one correct choice and is sent in its 5 cyclic permutations. The table shows the mean probability of the highest copy minus that of the lowest copy, in percentage points (\emph{Gap}). Equal probabilities give 0.}
\label{tab:copies}
\centering
\footnotesize
\setlength{\tabcolsep}{4pt}
\begin{tabular}{lrrr}
\toprule
 & \multicolumn{1}{c}{MetaTool} & \multicolumn{1}{c}{HotpotQA} & \multicolumn{1}{c}{ECtHR} \\
\cmidrule(lr){2-2} \cmidrule(lr){3-3} \cmidrule(lr){4-4}
Model & Gap & Gap & Gap \\
\midrule
Jev & 18 & 26 & 17 \\
Liquid d1 & 77 & 69 & 37 \\
GPT-6 Luna Decisions & 79 & 75 & 51 \\
\bottomrule
\end{tabular}

\end{table}

\begin{table}[t]
\caption{The share of answers that predict the copy in the first and in the last position, on the lists of Table~\ref{tab:copies}. An equal share is 20\%.}
\label{tab:positions}
\centering
\resizebox{\linewidth}{!}{\begin{tabular}{lrrrrrr}
\toprule
 & \multicolumn{2}{c}{MetaTool} & \multicolumn{2}{c}{HotpotQA} & \multicolumn{2}{c}{ECtHR} \\
\cmidrule(lr){2-3} \cmidrule(lr){4-5} \cmidrule(lr){6-7}
Model & First & Last & First & Last & First & Last \\
\midrule
Jev & 36\% & 25\% & 66\% & 16\% & 10\% & 56\% \\
Liquid d1 & 98\% & 1\% & 98\% & 1\% & 97\% & 1\% \\
GPT-6 Luna Decisions & 98\% & 1\% & 98\% & 1\% & 76\% & 1\% \\
\bottomrule
\end{tabular}
}
\end{table}

\begin{table}[t]
\caption{The highest exact set accuracy over the thresholds 0, 0.01, \ldots, 1. Exact set accuracy is the share of examples in which every correct choice is at or above the threshold and every distractor is below it. \emph{List} uses the probabilities of the list and \emph{Yes/no} the probability of yes of each choice.}
\label{tab:thresholds}
\centering
\resizebox{\linewidth}{!}{\begin{tabular}{lrrrrrr}
\toprule
 & \multicolumn{2}{c}{MetaTool} & \multicolumn{2}{c}{HotpotQA} & \multicolumn{2}{c}{ECtHR} \\
\cmidrule(lr){2-3} \cmidrule(lr){4-5} \cmidrule(lr){6-7}
Model & List & Yes/no & List & Yes/no & List & Yes/no \\
\midrule
Jev & 0.57 & 0.31 & 0.62 & 0.29 & 0.52 & 0.44 \\
Liquid d1 & 0.46 & 0.38 & 0.64 & 0.27 & 0.48 & 0.54 \\
GPT-6 Luna Decisions & 0.34 & 0.24 & 0.49 & 0.27 & 0.54 & 0.38 \\
\bottomrule
\end{tabular}
}
\end{table}

\paragraph{Correct choices receive unequal probabilities.}
When a list has two or more correct choices, the mean probability of the highest correct choice exceeds that of the lowest by 55 to 74 percentage points (Table~\ref{tab:allocation}, with 95\% intervals in Appendix~\ref{app:intervals}). A correct choice receives less than $1/n$ in 36\% to 60\% of the lists. The models rank every correct choice above every distractor in 52\% to 85\% of the lists. In 28\% to 50\% of those lists, a correct choice receives less than $1/n$. When the highest-probability correct choice is removed, the mean renormalized probability of the next correct choice differs from the mean probability the model assigns to that choice after the removal by 0.01 to 0.13 across models and datasets (Appendix~\ref{app:tables}). Correct choices are not equally relevant, and a probability below $1/n$ by itself is not a wrong allocation. However, relevance alone does not explain the size of the gap, as further demonstrated by the calibration and identical choices experiments below.

\paragraph{Identical copies receive unequal probabilities.}
In lists of five identical copies of a correct choice, the mean probability of the highest copy exceeds that of the lowest by 17 to 79 percentage points (Table~\ref{tab:copies}). Similar to \citet{deusser2026evaluating}, we observe that Jev and Liquid d1 show little position bias on lists with one correct tool, with accuracy changing by at most 4 percentage points between the first and the tenth place (Appendix~\ref{app:position}). However, identical copies of a correct choice still receive unequal probabilities. For Liquid d1 and GPT-6 Luna Decisions the gap between the highest and the lowest copy is 37 to 79 percentage points, and they predict the copy in the first position in 76\% to 98\% of their answers (Table~\ref{tab:positions}). For Jev it is 17 to 26 percentage points, and on ECtHR it predicts the copy in the last position in 56\% of its answers. Table~\ref{tab:copies-one} shows one HotpotQA list, in which Liquid d1 and GPT-6 Luna Decisions give 0.66 and 0.82 to the first copy. A choice can therefore receive a low probability for reasons unrelated to its content.

\paragraph{Thresholding.}
To give threshold-based selection the strongest possible advantage, we report the maximum exact set accuracy over all evaluated thresholds on each dataset. Even under this optimistic setting, exact set accuracy ranges from 0.34 to 0.64, despite the models ranking all correct choices above all distractors in 52\% to 85\% of the examples. For our lists of 10 choices each, the thresholds that maximize exact set accuracy range from 0.02 to 0.08 (Table~\ref{tab:thresholds}, Figure~\ref{fig:exact-set}). Thresholds this low have no practical interpretation, as they select choices to which the model gives a probability of a few percent. We repeat this process for the yes/no questions. The models rank all correct choices above all distractors in 42\% to 76\% of the examples, and the thresholds that maximize exact set accuracy range from 0.05 to 0.63. The highest, 0.63, is for Liquid d1 on ECtHR, where exact set accuracy is 0.54.

\paragraph{Calibration.}
When choice lists contain a single correct choice, the ECE is 1 to 14 percentage points. However, when lists with multiple correct choices are presented as yes/no questions, the probability of yes departs from the observed frequency in intermediate bins. While the ECE over all bins is 3 to 9 percentage points, it is influenced by low probabilities. Between 46\% and 78\% of the answers fall in the bin from 0 to 0.1, and at least 92\% of them are about distractors. Once this bin is removed, the ECE rises to 8 to 18 percentage points (Table~\ref{tab:calibration}, Figure~\ref{fig:reliability}). On intermediate bins, the gap between the observed frequency and the mean probability of yes reaches 19 percentage points for Jev and 40 percentage points on ECtHR. The probability of yes is therefore not calibrated with the observed frequency of correct choices.

\paragraph{No correct choice.}
When no choice in a list is correct, Jev and Liquid d1 answer every list, with a mean confidence of 0.49 to 0.87 (Appendix~\ref{app:abstention}).

\FloatBarrier

\section{Conclusion}

When a list has several correct choices, decision models concentrate probability on one of them, even when every correct choice ranks above every distractor. Identical copies of a correct choice receive different probabilities and show model- and dataset-dependent position bias. Even with thresholds optimized on the evaluation data, exact set accuracy remains below the share of examples in which the models rank every correct choice above every distractor. Asking separate yes/no questions for each choice lowers this accuracy for 8 of the 9 models and datasets, and the resulting intermediate probabilities are miscalibrated. Choice probabilities can rank choices, but they should not be interpreted as the probability that each choice is correct.

\section*{Limitations}
We study three decision models, Jev, Liquid d1 and GPT-6 Luna Decisions, through their hosted APIs. Generative language models that select a choice in text are outside the scope of this study. We use three English datasets, MetaTool, HotpotQA and ECtHR, and take the labels of each dataset as the set of correct choices. We study selection by one threshold for all examples of a dataset and by one yes/no question per choice. Other tasks with several correct choices, such as the diagnoses supported by a medical record or the vulnerability types present in code, and other selection rules, such as a threshold per example or a fixed number of selected choices, are left to future work.

\bibliography{references}

\appendix

\raggedbottom
\section{Passage choices in the opening example}
\label{app:example}
Table~\ref{tab:example-passages} gives the request of the introduction. Rubin's and Sharapova's passages contain the birth dates needed to answer which player is older.

\begin{table}[H]
\caption{The opening HotpotQA question, the instruction to choose every passage that supports the answer, and the 10 passage titles and texts in the order sent.}
\label{tab:example-passages}
\centering
\small
\begin{tabular}{@{}p{0.25\linewidth}p{0.68\linewidth}@{}}
\toprule
Question & who is older, Maria Sharapova or Chanda Rubin? \\
Instruction & Choose every passage in the list that supports the answer to the question. \\
\midrule
Title & Passage \\
\midrule
1997 IGA Classic -- Doubles & Chanda Rubin and Brenda Schultz-McCarthy were the defending champions but did not compete that year. \\
1997 EA-Generali Ladies Linz & The 1997 EA-Generali Ladies Linz was a women's tennis tournament played on indoor hard courts at the Intersport Arena in Linz in Austria that was part of Tier III of the 1997 WTA Tour.  \ldots \\
1997 Australian Open -- Women's Doubles & Chanda Rubin and Arantxa S\'{a}nchez Vicario were the defending champions, but competed this year with different partners.  \ldots \\
1997 State Farm Evert Cup -- Doubles & Chanda Rubin and Brenda Schultz-McCarthy were the defending champions but only Rubin competed that year with Mary Joe Fernandez. \\
Chanda Rubin & Chanda Rubin (born February 18, 1976) is an American tennis player.  \ldots \\
Maria Sharapova & Maria Yuryevna Sharapova (\ldots{} born April 19, 1987) is a Russian professional tennis player.  \ldots \\
1997 Bausch \& Lomb Championships -- Doubles & Chanda Rubin and Arantxa S\'{a}nchez Vicario were the defending champions but they competed with different partners that year, Rubin with Brenda Schultz-McCarthy and S\'{a}nchez Vicario with Larisa Savchenko. \\
2001 Bank of the West Classic -- Doubles & Chanda Rubin and Sandrine Testud were the defending champions, but none competed this year.  \ldots \\
1995 Schweppes Tasmanian International -- Doubles & Linda Harvey-Wild and Chanda Rubin were the defending champions but they competed with different partners that year, Harvey-Wild with Leila Meskhi and Rubin with Kristie Boogert. \\
2002 Wimbledon Championships -- Women's Doubles & Lisa Raymond and Rennae Stubbs were the defending champions, but lost in the quarterfinals to Anna Kournikova and Chanda Rubin. \\
\bottomrule
\end{tabular}

\end{table}

We sent the same question and passages with three Choice instructions. They were ``Choose every passage in the list that supports the answer to the question'' (\emph{Supports}), ``\ldots{} that is needed to answer the question'' (\emph{Needed}) and ``\ldots{} that has information that answers part of the question'' (\emph{Part of the question}). The matching yes/no questions were ``Does this passage support the answer to the question?'', ``Is this passage needed to answer the question?'' and ``Does this passage have information that answers part of the question?'', each followed by ``Passage:'' and the title and text of one passage.

\begin{table}[H]
\caption{Mean Choice probability. \emph{Distractor} is the highest probability among the other eight passages. Luna is GPT-6 Luna Decisions.}
\label{tab:example-wordings}
\centering
\resizebox{\linewidth}{!}{%
\begin{tabular}{llrrr}
\toprule
Instruction & Model & Rubin & Sharapova & Distractor \\
\midrule
Supports & Jev & 0.93 & 0.07 & 0.00 \\
 & Liquid d1 & 0.94 & 0.06 & 0.00 \\
 & Luna & 0.93 & 0.02 & 0.01 \\
\midrule
Needed & Jev & 0.78 & 0.22 & 0.00 \\
 & Liquid d1 & 0.91 & 0.09 & 0.00 \\
 & Luna & 0.95 & 0.01 & 0.01 \\
\midrule
Part of the question & Jev & 0.77 & 0.23 & 0.00 \\
 & Liquid d1 & 0.95 & 0.05 & 0.00 \\
 & Luna & 0.87 & 0.01 & 0.03 \\
\bottomrule
\end{tabular}}
\end{table}

\begin{table}[H]
\caption{Mean probability of yes, with \emph{Distractor} as in Table~\ref{tab:example-wordings}.}
\label{tab:example-wordings-yes}
\centering
\resizebox{\linewidth}{!}{%
\begin{tabular}{llrrr}
\toprule
Instruction & Model & Rubin & Sharapova & Distractor \\
\midrule
Supports & Jev & 0.31 & 0.14 & 0.05 \\
 & Liquid d1 & 0.16 & 0.07 & 0.05 \\
 & Luna & 0.47 & 0.00 & 0.01 \\
\midrule
Needed & Jev & 0.61 & 0.40 & 0.05 \\
 & Liquid d1 & 0.53 & 0.41 & 0.03 \\
 & Luna & 1.00 & 0.87 & 0.80 \\
\midrule
Part of the question & Jev & 0.63 & 0.52 & 0.10 \\
 & Liquid d1 & 0.81 & 0.52 & 0.08 \\
 & Luna & 1.00 & 0.96 & 0.50 \\
\bottomrule
\end{tabular}}
\end{table}

Under every instruction, each model gives Rubin's passage 0.77 or more and Sharapova's 0.23 or less. Sharapova's passage gets a probability of yes of 0.00 to 0.14 under Supports and 0.52 to 0.96 under Part of the question. The three instructions differ in meaning, and the comparison does not separate wording from meaning.

Table~\ref{tab:example-invented-passages} gives the same list with invented names, tournaments, dates and locations, sent with the Supports instruction and a question about the older and about the younger player. John is older than Jane. Under the older question, Jev and Liquid d1 favor John's passage and GPT-6 Luna Decisions favors Jane's. Under the younger question, all three favor Jane's (Table~\ref{tab:example-invented}).

\begin{table}[H]
\caption{The questions and passages with invented names. John stands for Chanda Rubin and Jane for Maria Sharapova, and the passages are in the order of Table~\ref{tab:example-passages}.}
\label{tab:example-invented-passages}
\centering
\small
\begin{tabular}{@{}p{0.25\linewidth}p{0.68\linewidth}@{}}
\toprule
Question & who is older, Jane or John? \\
Question & who is younger, Jane or John? \\
Instruction & Choose every passage in the list that supports the answer to the question. \\
\midrule
Title & Passage \\
\midrule
1998 Harlow Classic -- Doubles & John and Petra Lindqvist were the defending champions but did not compete that year. \\
1998 Varden Indoor Open & The 1998 Varden Indoor Open was a tennis tournament played on indoor hard courts at the Kestrel Arena in Varden in Switzerland that was part of Tier III of the 1998 Meridian Tour.  \ldots \\
1998 Corvale Open -- Doubles & John and Ilse Marquart were the defending champions, but competed this year with different partners.  \ldots \\
1998 Lakemont Cup -- Doubles & John and Petra Lindqvist were the defending champions but only John competed that year with Nora Castellan. \\
John & John (born December 12, 1976) is an Italian tennis player.  \ldots \\
Jane & Jane (born January 13, 1990) is an Australian professional tennis player.  \ldots \\
1998 Stonebridge Championships -- Doubles & John and Ilse Marquart were the defending champions but they competed with different partners that year, John with Petra Lindqvist and Marquart with Vera Kolesnik. \\
2003 Bay Harbor Classic -- Doubles & John and Camille Roussel were the defending champions, but none competed this year.  \ldots \\
1996 Redcliff International -- Doubles & Marta Hale-Winter and John were the defending champions but they competed with different partners that year, Hale-Winter with Sofia Nemec and John with Annika Brandt. \\
2004 Ashgrove Championships -- Doubles & Laura Pemberton and Renee Stokes were the defending champions, but lost in the quarterfinals to Irina Voss and John. \\
\bottomrule
\end{tabular}

\end{table}

\begin{table}[H]
\caption{Choice probability of John's and Jane's passages. Luna is GPT-6 Luna Decisions.}
\label{tab:example-invented}
\centering
\small
\begin{tabular}{llrrr}
\toprule
Question & Passage & Jev & Liquid d1 & Luna \\
\midrule
Who is older?   & John & 0.91 & 0.94 & 0.06 \\
                & Jane & 0.09 & 0.06 & 0.79 \\
Who is younger? & John & 0.02 & 0.15 & 0.03 \\
                & Jane & 0.98 & 0.85 & 0.89 \\
\bottomrule
\end{tabular}
\end{table}

\section{Removal of the highest correct choice}
\label{app:tables}
For each list with two or more correct choices, we sent the list again without the correct choice with the highest probability, and report the correct choice with the second highest probability. If the other choices kept their ratios, its probability after the removal would be its probability in the full list divided by one minus the probability of the removed choice (\emph{Renorm.}).

\begin{table}[H]
\caption{Mean probability of the second correct choice in the full list (\emph{Before}) and after the removal (\emph{After}).}
\label{tab:removal}
\centering
\resizebox{\linewidth}{!}{\begin{tabular}{lrrrrrr}
\toprule
 & \multicolumn{2}{c}{MetaTool} & \multicolumn{2}{c}{HotpotQA} & \multicolumn{2}{c}{ECtHR} \\
\cmidrule(lr){2-3} \cmidrule(lr){4-5} \cmidrule(lr){6-7}
Model & Before & After & Before & After & Before & After \\
\midrule
Jev & 0.11 & 0.71 & 0.17 & 0.66 & 0.19 & 0.71 \\
Liquid d1 & 0.16 & 0.62 & 0.19 & 0.68 & 0.16 & 0.68 \\
GPT-6 Luna Decisions & 0.11 & 0.53 & 0.16 & 0.60 & 0.15 & 0.68 \\
\bottomrule
\end{tabular}
}
\end{table}

\begin{table}[H]
\caption{Mean probability of the second correct choice after renormalisation (\emph{Renorm.}) and after the removal (\emph{After$^*$}), on the examples in which the choices other than the removed one hold 0.1 or more.}
\label{tab:removal-renormalised}
\centering
\resizebox{\linewidth}{!}{\begin{tabular}{lrrrrrr}
\toprule
 & \multicolumn{2}{c}{MetaTool} & \multicolumn{2}{c}{HotpotQA} & \multicolumn{2}{c}{ECtHR} \\
\cmidrule(lr){2-3} \cmidrule(lr){4-5} \cmidrule(lr){6-7}
Model & Renorm. & After$^*$ & Renorm. & After$^*$ & Renorm. & After$^*$ \\
\midrule
Jev & 0.77 & 0.72 & 0.80 & 0.67 & 0.72 & 0.71 \\
Liquid d1 & 0.64 & 0.61 & 0.80 & 0.70 & 0.70 & 0.67 \\
GPT-6 Luna Decisions & 0.40 & 0.49 & 0.61 & 0.60 & 0.59 & 0.68 \\
\bottomrule
\end{tabular}
}
\end{table}

\begin{figure*}[t]
\centering
\includegraphics[width=\linewidth]{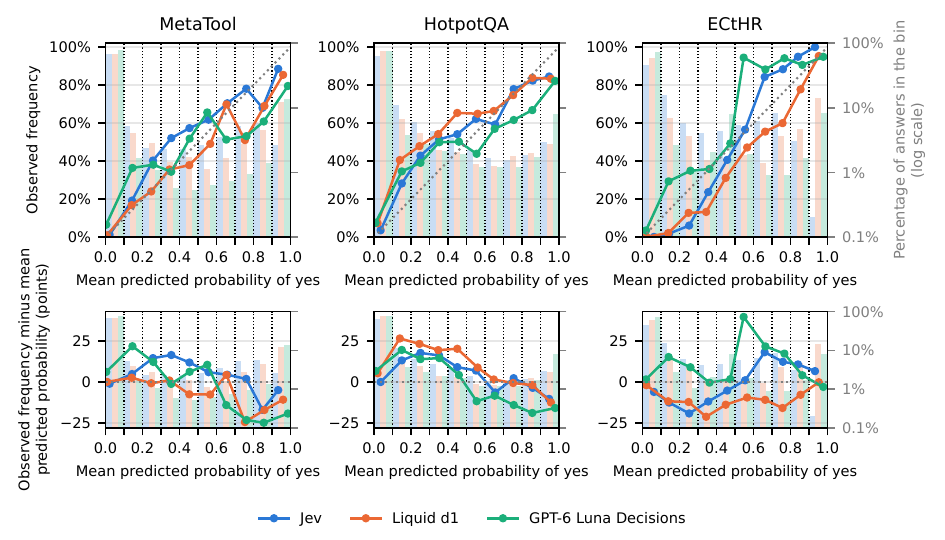}
\caption{Reliability diagram of the yes/no questions on examples with two or more correct choices, in bins of width 0.1. Each point is at the mean probability of yes of its bin. The top row gives the observed frequency of correct choices, and the bottom row the observed frequency minus the mean probability, in percentage points. The bars give the share of the answers in each bin, on a log scale.}
\label{fig:reliability}
\end{figure*}

\section{One list of identical copies}
\label{app:lists}
For the HotpotQA question ``What government position was held by the woman who portrayed Corliss Archer in the film Kiss and Tell?'', we sent a list of five identical copies of the Shirley Temple passage, each named by a random ID.

\begin{table}[H]
\caption{Probability of each copy in one run, in the order sent. The highest and the lowest probability of each model are in bold, and \emph{Gap} is the highest minus the lowest, in percentage points.}
\label{tab:copies-one}
\centering
\resizebox{\linewidth}{!}{%
\begin{tabular}{@{}rlrrr@{}}
\toprule
Position & ID & Jev & Liquid d1 & GPT-6 Luna Decisions \\
\midrule
1 & \texttt{7bg8} & \textbf{0.43} & \textbf{0.66} & \textbf{0.82} \\
2 & \texttt{c9jq} & 0.11 & 0.08 & 0.04 \\
3 & \texttt{3nwa} & 0.15 & \textbf{0.06} & 0.03 \\
4 & \texttt{6hp7} & \textbf{0.09} & 0.07 & \textbf{0.02} \\
5 & \texttt{2haw} & 0.22 & 0.13 & 0.09 \\
\midrule
 & Gap & 34 & 60 & 80 \\
\bottomrule
\end{tabular}}
\end{table}

\section{Calibration}
\label{app:calibration}
On lists with one correct choice, the ECE of the confidence is 8 to 14 percentage points on MetaTool and 1 to 3 on ECtHR (Table~\ref{tab:calibration-one}). Without the bin from 0 to 0.1, the ECE of the yes/no questions is highest for GPT-6 Luna Decisions on MetaTool, 18 percentage points, and lowest for Liquid d1 and GPT-6 Luna Decisions on ECtHR, 8 (Table~\ref{tab:calibration}).

\begin{table}[H]
\caption{ECE in percentage points of the confidence on lists with one correct choice.}
\label{tab:calibration-one}
\centering
\footnotesize
\begin{tabular}{lrr}
\toprule
 & \multicolumn{2}{c}{One correct choice} \\
\cmidrule(lr){2-3}
 & \multicolumn{1}{c}{MetaTool} & \multicolumn{1}{c}{ECtHR} \\
\cmidrule(lr){2-2} \cmidrule(lr){3-3}
Model & ECE & ECE \\
\midrule
Jev & 8 & 3 \\
Liquid d1 & 9 & 1 \\
GPT-6 Luna Decisions & 14 & 1 \\
\bottomrule
\end{tabular}

\end{table}

\begin{table}[H]
\caption{ECE in percentage points of the probability of yes on the yes/no questions, over all bins (\emph{All bins}) and without the bin from 0 to 0.1 (\emph{From 0.1}).}
\label{tab:calibration}
\centering
\resizebox{\linewidth}{!}{\begin{tabular}{lrrrrrr}
\toprule
 & \multicolumn{6}{c}{Yes/no questions} \\
\cmidrule(lr){2-7}
 & \multicolumn{2}{c}{MetaTool} & \multicolumn{2}{c}{HotpotQA} & \multicolumn{2}{c}{ECtHR} \\
\cmidrule(lr){2-3} \cmidrule(lr){4-5} \cmidrule(lr){6-7}
Model & All bins & From 0.1 & All bins & From 0.1 & All bins & From 0.1 \\
\midrule
Jev & 3 & 9 & 4 & 12 & 9 & 12 \\
Liquid d1 & 3 & 10 & 8 & 17 & 4 & 8 \\
GPT-6 Luna Decisions & 9 & 18 & 9 & 15 & 3 & 8 \\
\bottomrule
\end{tabular}
}
\end{table}

\section{Lists with no correct choice}
\label{app:abstention}
We build the lists with no correct choice from examples with any number of correct choices and leave out every correct choice. A list holds either random choices from the dataset that are not in the example's list, or the example's own distractors. We send each list without and with a ``none of the above'' choice. Without it, we report the confidence of the prediction. With it, ``none of the above'' is the only correct answer, and we report the share of predictions that select it.

Without a ``none of the above'' choice, the mean confidence is 0.49 to 0.71 on HotpotQA and 0.74 to 0.87 on MetaTool and ECtHR (Table~\ref{tab:abstention}). With it, the models select it in 93\% to 100\% of the lists of random choices and in 79\% to 88\% of the HotpotQA lists of the example's own distractors, and in 7\% to 37\% of the ECtHR lists (Table~\ref{tab:nota}).

\begin{figure*}[t]
\centering
\includegraphics[width=\linewidth]{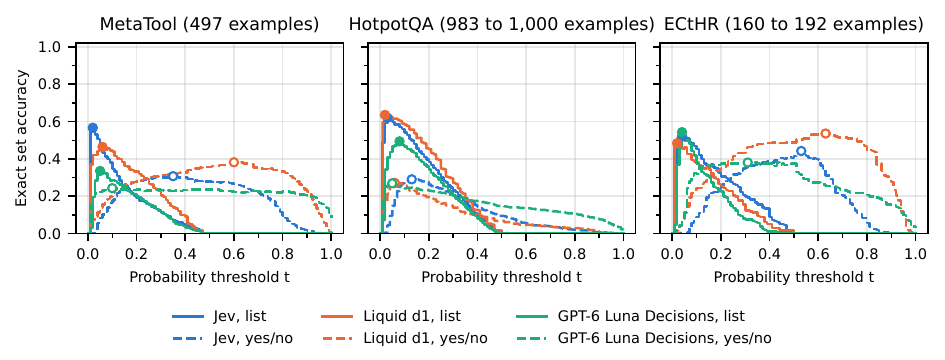}
\caption{Exact set accuracy, the share of examples in which every correct choice is at or above the threshold $t$ and every distractor is below it. Above 0.50 the list curves are 0 by construction.}
\label{fig:exact-set}
\end{figure*}

\begin{table}[H]
\caption{Mean confidence on the answered lists with no correct choice and no ``none of the above'' choice (\emph{Conf.}). Appendix~\ref{app:refusals} counts the refused lists.}
\label{tab:abstention}
\centering
\resizebox{\linewidth}{!}{\begin{tabular}{lrrrrr}
\toprule
 & \multicolumn{2}{c}{Random choices} & \multicolumn{3}{c}{The example's own} \\
\cmidrule(lr){2-3} \cmidrule(lr){4-6}
 & \multicolumn{1}{c}{MetaTool} & \multicolumn{1}{c}{HotpotQA} & \multicolumn{1}{c}{MetaTool} & \multicolumn{1}{c}{HotpotQA} & \multicolumn{1}{c}{ECtHR} \\
\cmidrule(lr){2-2} \cmidrule(lr){3-3} \cmidrule(lr){4-4} \cmidrule(lr){5-5} \cmidrule(lr){6-6}
Model & Conf. & Conf. & Conf. & Conf. & Conf. \\
\midrule
Jev & 0.78 & 0.58 & 0.80 & 0.61 & 0.83 \\
Liquid d1 & 0.74 & 0.49 & 0.77 & 0.67 & 0.87 \\
GPT-6 Luna Decisions & 0.79 & 0.71 & 0.78 & 0.65 & 0.87 \\
\bottomrule
\end{tabular}
}
\end{table}

\begin{table}[H]
\caption{Share of predictions that select ``none of the above'', the only correct choice (\emph{Selected}).}
\label{tab:nota}
\centering
\resizebox{\linewidth}{!}{\begin{tabular}{lrrrr}
\toprule
 & \multicolumn{2}{c}{Random choices} & \multicolumn{2}{c}{The example's own} \\
\cmidrule(lr){2-3} \cmidrule(lr){4-5}
 & \multicolumn{1}{c}{MetaTool} & \multicolumn{1}{c}{HotpotQA} & \multicolumn{1}{c}{HotpotQA} & \multicolumn{1}{c}{ECtHR} \\
\cmidrule(lr){2-2} \cmidrule(lr){3-3} \cmidrule(lr){4-4} \cmidrule(lr){5-5}
Model & Selected & Selected & Selected & Selected \\
\midrule
Jev & 95\% & 100\% & 84\% & 37\% \\
Liquid d1 & 98\% & 100\% & 88\% & 7\% \\
GPT-6 Luna Decisions & 93\% & 100\% & 79\% & 17\% \\
\bottomrule
\end{tabular}
}
\end{table}

\section{Position of the correct tool}
\label{app:position}
For 2,795 MetaTool queries with one correct tool, from the test files \texttt{similar\_tools} and \texttt{scenario}, we placed the correct tool first or tenth in a list of 10 tools that ends with the choice ``None''. The other nine tools keep their order. GPT-6 Luna Decisions refused 2 of the lists.

\begin{table}[H]
\caption{Accuracy with the correct tool first and tenth, and the change in percentage points.}
\label{tab:position}
\centering
\small
\begin{tabular}{lrrr}
\toprule
Model & First & Tenth & Change \\
\midrule
Jev & 0.78 & 0.80 & +2 \\
Liquid d1 & 0.69 & 0.74 & +4 \\
GPT-6 Luna Decisions & 0.80 & 0.67 & $-13$ \\
\bottomrule
\end{tabular}
\end{table}

\section{Runs}
\label{app:runs}
Each model received 30,457 requests. The lists of identical copies took 5,139 requests per model, and the yes/no questions 497 on MetaTool, 1,000 on HotpotQA and 1,000 on ECtHR.

\begin{table}[H]
\caption{Examples by dataset and list type, with the number of lists or yes/no questions per model in parentheses where it differs. GPT-6 Luna Decisions refused some of them (Appendix~\ref{app:refusals}).}
\label{tab:runs}
\centering
\small
\resizebox{\linewidth}{!}{%
\begin{tabular}{lrrr}
\toprule
List & MetaTool & HotpotQA & ECtHR \\
\midrule
Several correct choices & 497 & 7,405 & 192 (1,920) \\
Five identical copies & 3,292 (16,460) & 1,000 (5,000) & 847 (4,235) \\
One correct choice & 2,795 (16,770) & -- & 655 (6,550) \\
No correct choice, random choices & 3,292 (16,460) & 1,000 (5,000) & -- \\
No correct choice, the example's own & 995 (3,980) & 7,377 & 153 (1,530) \\
Random choices and ``none of the above'' & 995 (4,975) & 1,000 (5,000) & -- \\
The example's own and ``none of the above'' & -- & 1,000 (5,000) & 153 \\
Highest correct choice removed & 497 & 7,384 & 192 \\
Yes/no question per choice & 497 (4,970) & 1,000 (10,000) & 1,000 (10,000) \\
Selecting by a threshold & 497 & 1,000 & 192 \\
\bottomrule
\end{tabular}}
\end{table}

\section{Selecting by a threshold}
\label{app:thresholds}
For a yes/no question, a probability of yes below 0.5 is an answer of no. The thresholds that maximize the exact set accuracy of the yes/no questions are below 0.5 for 6 of the 9 models and datasets, and 0.13 or lower on HotpotQA. At 0.5, the exact set accuracy of the yes/no questions is 0.07 to 0.49 (Figure~\ref{fig:exact-set}).

\begin{table}[H]
\caption{The threshold with the highest exact set accuracy on the examples of Table~\ref{tab:thresholds} (\emph{Threshold}), the share of examples with every correct choice above every distractor (\emph{Ranked}), and the share of those examples whose set is exact at that threshold (\emph{Exact among ranked}), precision and recall at that threshold, and the highest F1 over all thresholds, with its threshold in parentheses (\emph{Best F1}). Precision, recall and F1 pool the choices of all examples. \emph{List} uses the probabilities of the list, and \emph{Yes/no} the probability of yes of each choice. The highest and the lowest value of each setting in the columns Threshold and Exact among ranked are in bold.}
\label{tab:threshold-detail}
\centering
\resizebox{\linewidth}{!}{\begin{tabular}{lllrrrrrr}
\toprule
Model & Dataset & Setting & Threshold & Ranked & Exact among ranked & Precision & Recall & Best F1 \\
\midrule
Jev & MetaTool & List & \textbf{0.02} & 0.73 & \textbf{0.78} & 0.86 & 0.87 & 0.87 (0.02) \\
 & MetaTool & Yes/no & 0.35 & 0.52 & 0.59 & 0.70 & 0.83 & 0.78 (0.25) \\
\midrule
Liquid d1 & MetaTool & List & 0.06 & 0.70 & 0.66 & 0.80 & 0.84 & 0.83 (0.05) \\
 & MetaTool & Yes/no & 0.60 & 0.61 & 0.63 & 0.76 & 0.81 & 0.79 (0.47) \\
\midrule
GPT-6 Luna Decisions & MetaTool & List & 0.05 & 0.52 & 0.65 & 0.73 & 0.77 & 0.75 (0.05) \\
 & MetaTool & Yes/no & 0.10 & 0.42 & 0.57 & 0.68 & 0.74 & 0.71 (0.12) \\
\midrule
Jev & HotpotQA & List & 0.03 & 0.83 & 0.75 & 0.84 & 0.90 & 0.87 (0.04) \\
 & HotpotQA & Yes/no & 0.13 & 0.57 & 0.51 & 0.52 & 0.85 & 0.66 (0.19) \\
\midrule
Liquid d1 & HotpotQA & List & \textbf{0.02} & 0.85 & 0.74 & 0.82 & 0.93 & 0.88 (0.04) \\
 & HotpotQA & Yes/no & 0.06 & 0.55 & \textbf{0.49} & 0.53 & 0.82 & 0.65 (0.09) \\
\midrule
GPT-6 Luna Decisions & HotpotQA & List & \textbf{0.08} & 0.71 & 0.70 & 0.87 & 0.80 & 0.84 (0.08) \\
 & HotpotQA & Yes/no & \textbf{0.05} & 0.51 & 0.53 & 0.55 & 0.79 & 0.65 (0.13) \\
\midrule
Jev & ECtHR & List & 0.04 & 0.79 & 0.66 & 0.84 & 0.89 & 0.87 (0.03) \\
 & ECtHR & Yes/no & 0.53 & 0.70 & 0.63 & 0.83 & 0.82 & 0.83 (0.45) \\
\midrule
Liquid d1 & ECtHR & List & \textbf{0.02} & 0.76 & \textbf{0.64} & 0.79 & 0.91 & 0.85 (0.01) \\
 & ECtHR & Yes/no & \textbf{0.63} & 0.76 & \textbf{0.71} & 0.85 & 0.89 & 0.87 (0.57) \\
\midrule
GPT-6 Luna Decisions & ECtHR & List & 0.04 & 0.72 & 0.76 & 0.85 & 0.85 & 0.85 (0.04) \\
 & ECtHR & Yes/no & 0.31 & 0.64 & 0.60 & 0.78 & 0.81 & 0.80 (0.18) \\
\bottomrule
\end{tabular}
}
\end{table}

\section{Refusals of GPT-6 Luna Decisions}
\label{app:refusals}
GPT-6 Luna Decisions refused 19,724 of the 168,322 lists and yes/no questions in its 30,457 requests (11.7\%). It refused 14,420 of 31,435 lists of random choices, 1,374 of 25,695 lists of identical copies, 3,792 of 86,222 other lists, and 138 of 24,970 yes/no questions. A refusal has no probabilities and no reason. We sent one refused request again seven times, with one change each time. The model answered the same passages when the query was changed to one that a passage fits, and the same query when a ``none of the above'' choice or a fitting passage was added.

\section{Confidence intervals}
\label{app:intervals}
The intervals are 95\% bootstrap intervals over the examples, from 2,000 resamples. A resample keeps every answer to an example together. The exact set accuracy is read at the threshold with the highest exact set accuracy on all the examples, and its interval leaves out the choice of that threshold.

\begin{table}[H]
\caption{The figures of Tables~\ref{tab:allocation}, \ref{tab:copies} and~\ref{tab:thresholds} with their 95\% intervals. \emph{Gap} and \emph{Gap of the copies} are in percentage points, and \emph{Ranked} and \emph{Below $1/n$} in percent.}
\label{tab:intervals}
\centering
\resizebox{\linewidth}{!}{\begin{tabular}{lrrr}
\toprule
Figure & Jev & Liquid d1 & GPT-6 Luna Decisions \\
\midrule
\multicolumn{4}{l}{\emph{MetaTool}} \\
Gap & 74 [71, 76] & 55 [52, 57] & 55 [53, 58] \\
Ranked & 73 [69, 77] & 70 [66, 74] & 52 [47, 56] \\
Below $1/n$ & 50 [45, 55] & 29 [24, 34] & 35 [30, 41] \\
Gap of the copies & 18 [18, 18] & 77 [76, 77] & 79 [78, 79] \\
Exact set, list & 0.57 [0.52, 0.61] & 0.46 [0.42, 0.51] & 0.34 [0.29, 0.38] \\
Exact set, yes/no & 0.31 [0.27, 0.35] & 0.38 [0.34, 0.42] & 0.24 [0.21, 0.28] \\
\midrule
\multicolumn{4}{l}{\emph{HotpotQA}} \\
Gap & 60 [59, 60] & 55 [54, 56] & 57 [56, 58] \\
Ranked & 82 [81, 82] & 85 [84, 86] & 70 [69, 71] \\
Below $1/n$ & 31 [30, 32] & 28 [27, 29] & 29 [28, 30] \\
Gap of the copies & 26 [25, 26] & 69 [68, 70] & 75 [74, 76] \\
Exact set, list & 0.62 [0.59, 0.65] & 0.64 [0.61, 0.67] & 0.49 [0.46, 0.53] \\
Exact set, yes/no & 0.29 [0.26, 0.32] & 0.27 [0.25, 0.30] & 0.27 [0.24, 0.30] \\
\midrule
\multicolumn{4}{l}{\emph{ECtHR}} \\
Gap & 58 [54, 62] & 57 [53, 61] & 61 [57, 64] \\
Ranked & 79 [72, 84] & 76 [69, 81] & 73 [66, 79] \\
Below $1/n$ & 36 [29, 45] & 42 [34, 50] & 45 [37, 53] \\
Gap of the copies & 17 [17, 17] & 37 [36, 38] & 51 [50, 51] \\
Exact set, list & 0.52 [0.45, 0.59] & 0.48 [0.41, 0.56] & 0.54 [0.47, 0.62] \\
Exact set, yes/no & 0.44 [0.38, 0.51] & 0.54 [0.46, 0.61] & 0.38 [0.30, 0.46] \\
\bottomrule
\end{tabular}
}
\end{table}

\section{Code and data availability}
\label{app:availability}
The code and the instructions sent to the models are at \url{https://github.com/Rushab1/gr-jev} under the MIT License. This paper uses the version tagged \texttt{arxiv-v1}. The saved model responses are not in the repository.

\end{document}